\documentclass[graybox]{svmult}

\usepackage{type1cm}        
\usepackage{makeidx}         
\usepackage{graphicx}        
\usepackage{multicol}        
\usepackage[bottom]{footmisc}

\usepackage{newtxtext}       %
\usepackage{newtxmath}       

\makeindex             

\usepackage{amssymb}  
\usepackage{mathrsfs} 
\usepackage{stmaryrd} 

\usepackage{color}  

\usepackage{enumerate}

\usepackage{appendix} 
\usepackage{cleveref}

\newcommand{\EEE}{{\mathbb{E}}}
\newcommand{\R}{{\mathbb{R}}}

\newcommand{\A}{{\mathscr{A}}}
\newcommand{\BB}{{\mathscr{B}}}
\newcommand{\C}{{\mathscr{C}}}
\newcommand{\DDD}{{\mathscr{D}}}

\newcommand{\eq}[1]{\begin{align}#1\end{align}}

\newcommand{\blue}[1]{{{#1}}}
\newcommand{\red}[1]{{{#1}}}

\begin{document}


\title*{What is important about the No Free Lunch theorems?}
\author{David H. Wolpert}
\institute{David Wolpert\at Santa Fe Institute, New Mexico, USA, \email{david.h.wolpert@gmail.com}
}
%
%
\maketitle

\abstract{The No Free Lunch theorems prove that under a uniform distribution over induction problems (search problems or learning problems), 
all induction algorithms perform equally. As I discuss in this chapter, the importance of the theorems arises by using them to analyze scenarios  
involving \textit{non-uniform} distributions, and to compare different
algorithms, without any assumption about the distribution over problems at all. In particular,
the theorems prove that \textit{anti}-cross-validation (choosing among a set of candidate algorithms based
on which has \textit{worst} out-of-sample behavior) performs as well as cross-validation, unless one
makes an assumption --- which has never been formalized --- about how
the distribution over induction problems, on the one hand, is related to the set of algorithms one is choosing among
using (anti-)cross validation, on the other. In addition,
they establish strong caveats concerning the significance of the many results in the literature which establish the strength
of a particular algorithm without assuming a particular distribution. They also 
motivate a ``dictionary'' between supervised learning and improve blackbox optimization, which allows one to ``translate'' techniques
from supervised learning into the domain of blackbox optimization, thereby strengthening blackbox optimization algorithms. In addition to these topics,
I also briefly discuss their implications for philosophy of science.
}


\section{Introduction}   

The first of what are now called the No Free Lunch (NFL) theorems were published in~\cite{wolp96ab}.  
Soon after publication they were popularized in~\cite{scha94}, building on a preprint version of~\cite{wolp96ab}. 
Those first theorems focused on (supervised) {machine} learning. Loosely
speaking, they can be viewed as  a formalization and elaboration of  informal concerns about the
legitimacy of inductive inference that date back to David Hume (if not earlier). Shortly
after these original theorems were published, additional NFL theorems that apply to search were introduced in~\cite{woma97}.
Broadly speaking, the NFL theorems say that under a uniform distribution over problems (be they supervised learning
problems or search problems), all algorithms perform equally.

The NFL theorems have stimulated a huge amount of research, with over 10,000 citations of~\cite{woma97}
alone by summer 2020, according to Google Scholar. Much of the early work focused
on finding other prior distributions over problems besides a uniform distribution that still result in all algorithms having the same expected performance~\cite{whitley2011no,igel2005no}.
Other, more recent work has extended NFL to other domains, beyond (classical physics based)
supervised learning~\cite{poland2020no}, and beyond learning and search entirely~\cite{peel2017ground}. 

However, as stated in~\cite{woma97}, perhaps the primary significance of the NFL theorems for search 
is what they tell us  about ``the underlying mathematical `skeleton' of optimization theory \emph{before the
`flesh' of the probability distributions of a particular context
and set of optimization problems are imposed}". So in particular, while
the NFL theorems have strong implications \emph{if} one believes in a uniform distribution
over optimization problems, in no sense should they be interpreted as advocating 
such a distribution. Rather such a distribution is used as a tool, to prove results
concerning \textit{non-uniform} distributions, and in addition to compare different
search algorithms, without any direct assumption about the distribution over problems at all.

In this chapter I describe these aspects of the NFL theorems. After presenting the inner product formula
that determines the performance of any search algorithm, I then expand on the inner product
formula to present the NFL theorems for search and supervised learning. As an example of the true
significance of the NFL theorems, I consider ``anti-cross-validation''
which is the meta-algorithm that chooses among a candidate set of algorithms based on which has the
\textit{worst} out-of-sample performance on a given data set. (In contrast, standard cross-validation chooses
the algorithm with the best such performance.) As I discuss, the NFL theorems mean that anti-cross-validation outperforms
cross-validation as often as vice-versa, over the set of all objective functions. So without making some assumption about the relationship
between the candidate algorithms and the distribution over optimization problems, one cannot
even justify using cross-validation.

Following up on this, I briefly discuss how the NFL theorems are consistent with the (very) many proofs in the literature
that provide lower bounds on the performance of particular algorithms without making any
assumptions about the distribution over problems that are fed to those algorithms. I also
point out the implications of the NFL theorems for the entire scientific enterprise, i.e., for 
philosophy of science~\cite{godfrey2009theory}.

I then discuss how the fact that there are NFL theorems for both search and for supervised
learning is symptomatic of the deep formal relationship between those two fields. Once
that relationship is disentangled, it suggests many ways that we can exploit practical
techniques that were first developed in supervised learning to help us do search. I
summarize some experiments that confirm the power of search algorithms developed
in this way. 

After this I briefly discuss the various \emph{free} lunch theorems that have been derived,
which establish \textit{a priori} benefits for using one algorithm rather than another.
I end by discussing possible directions for future research.

\section{The inner product at the heart of all search}

Let $X$ be a countable \textbf{search space}, and specify an \textbf{objective
function} $f : X \rightarrow Y$ where $Y \subset \R$ is a countable set. 
Sometimes  an objective function is instead called
a ``search problem", ``fitness function", ``cost function", etc.
Use the term \textbf{data set} to mean any set of $m$ separate
pairs $(x \in X, f(x))$, written as  $d^m = \{d_X^m, d_Y^m\}$. A \textbf{search algorithm} is a function $\A$ that maps any $d^m$
for any $m \in \{0, 1, \ldots \}$ to an $x \not \in d_X^m$. Examples range from simulated annealing
to genetic algorithms to hill-descending. By iteratively running a search algorithm to produce
a new sample point $x$ and then evaluating $f(x)$ we can build 
successively larger data sets: $d^{m+1} = d^m \cup (A(d_X^m), f[A(d_X^m)])$ for all $m \ge 0$.

Suppose we are given an arbitrary \textbf{performance measure} $\Phi : d_Y^m \rightarrow \R$. Then we can
evaluate how the performance of a given search algorithm on a given objective function changes
as it is run on that function. Note that to ``normalize'' different search
algorithms, we only consider their behavior in terms of generating new points at which to sample
the objective function that are not yet in the data set. (Equivalently, if an algorithm chooses a new point to
sample that is already in its data set, we allow it to ``try again''.) This is crucial; we are only interested in \textit{off-data-set}
behavior.

For simplicity, from now on I restrict attention to deterministic search algorithms and deterministic objective functions neither
of which varies from one iteration of the search algorithm to the next. However, everything presented in this
paper can be extended in a straightforward way to the case of a stochastic
search algorithm, stochastic objective function, time-varying objective function, etc.

%

In practice often one does not know $f$
explicitly. This is the case whenever $f$ is a ``blackbox", or an ``oracle", that one can
sample at a particular $x$, but does not know in closed form. Moreover, often even if a practitioner
does explicitly know $f$, they \emph{act} as though they
do not know it, for example when they choose what search algorithm to use on $f$. For example,
often someone trying to solve a particular instance of the Traveling Salesman Problem (TSP) will use the same search
algorithm that they would use on any other instance of TSP. In such a case, they are behaving exactly as they
would if they only knew that the objective function is an TSP, without knowing specifically which
one it is.

These kinds of uncertainty about the precise $f$ being searched can be expressed as a distribution $P(f)$. Say we are given
such a $P(f)$, \red{along with a search algorithm, and a real-valued measure of the performance
of that algorithm when it is run on any objective function $f$. Then we can solve for 
the probability that the
algorithm results in a performance value $\phi$. The result}
is an inner product of two real-valued vectors each indexed by $f$. \red{(See Appendix.)} The first
of those vectors gives all the details of how the search algorithm operates, but nothing concerning the world in which one
deploys that search algorithm. The second vector is $P(f)$. All the details of the world in
which one deploys that search algorithm are specified in this vector, but nothing
concerning the search algorithm itself.

\red{This result} tells us that at root, how well any search algorithm performs
is determined by how well it is  ``aligned'' with the distribution $P(f)$ that governs the problems
on which that algorithm is run.
For example, it
means that the (tens of?) thousand of person-years of research into the TSP
have (presumably) resulted in algorithms aligned
with the implicit $P(f)$ describing traveling salesman problems
of interest to TSP researchers.

\section{The No Free Lunch theorems for search}
\label{sec:nfl_search}

The inner product result governs how well any particular
search algorithm does in practice. Therefore, either explicitly or implicitly,
it serves as the basis for any practitioner who chooses a search algorithm
to use in a given scenario. More
precisely, the designer of any search algorithm
first specifies a $P(f)$ {\blue{(usually implicitly, e.g., by restricting attention to
a class of optimization problems)}}. Then they specify a performance
measure $\Phi$ (sometimes explicitly). {\blue{Properly speaking, they should
then solve for the search algorithm that the inner product result tells us will
have the best distribution of values of that performance measure, for that $P(f)$. In practice
though, instead informal arguments are often used to motivate the search algorithm.}}


In addition to \red{governing both how a practitioner should
design their search algorithm, and how well the actual algorithm they use
performs, the inner}
product result can be used to make more general statements
about search, \blue{results} that hold for all $P(f)$'s. It does this by allowing us to compare the
performance of a given search algorithm on different subsets of the set of all objective functions. 
\red{The result}
is the NFL theorem for search. It tells us that
if any search algorithm performs particularly well on one set of objective functions,
it must perform correspondingly \emph{poorly} on all other objective functions.

This implication is the primary significance of the NFL theorem for search. 
To illustrate it, choose the first set to be the set of objective functions
on which your favorite search algorithm performs better than the 
\emph{purely random search algorithm}, which chooses the next
sample point randomly. Then the NFL for search theorem says 
that compared to random search, your favorite search algorithm ``loses on as many'' objective functions  as it wins 
(if one weights wins / losses by the amount of the win / loss). This is true no matter what performance measure you use. 

As another example, say that your performance measure prefers low values of
the objective function \blue{to} high values\blue{, i.e., that your goal is to find low
values of the objective rather than high ones}. Then \red{we can use the NFL theorem
for search to compare}  a hill-descending algorithm 
to a hill-\emph{ascending} algorithm\red{, i.e., to} \blue{an algorithm that} ``tries'' to do as poorly as possible according
to the objective function. \red{The conclusion is}
that the hill-descending algorithm ``loses \blue{to the hill-ascending algorithm} on as
many'' objective functions as it wins. The lesson is that without arguing
for a particular $P(f)$ that is biased towards the objective functions on which one's 
favorite search algorithm performs well, one  has no formal justification that that algorithm 
has good performance.

A secondary implication of the NFL theorem for search is that \emph{if} it so happens that you assume / 
believe that $P(f)$ is uniform, then the \blue{average over $f$'s used in the NFL for search theorem
is the same as $P(f)$. In this case,}
you must conclude that all search algorithms perform equally well\red{ for your assumed $P(f)$}.
This conclusion is only as legitimate as is the assumption for $P(f)$ it is based on.
Once other $P(f)$'s are allowed, the conclusion need not hold.

An important
point in this regard is that simply allowing $P(f)$ to \red{be non-}uniform, \emph{by itself}, does not 
invalidate the NFL theorem for search. \blue{Arguments that $P(f)$ is non-uniform in the real world
do not, by themselves, establish anything whatsoever about what search algorithm to use in 
the real world. 

In fact,  allowing $P(f)$'s to vary provides us with a new NFL theorem. In this
new theorem, rather than compare the performance of two search algorithms over all $f$'s, we
compare them over all $P(f)$'s. The result is what one might expect: If any given search algorithm
performs better than another over a given set of $P(f)$'s, then it must perform corresponding worse
on all other $P(f)$'s. (See appendix for proof.)}

\section{The supervised learning No Free Lunch theorems}

The discussion \red{above tells us} that if we only knew \blue{and properly exploited}
$P(f)$, we would be able to design an associated search algorithm
that performs better than random. This suggests that we try to \blue{use a search
process itself to}
learn something about \blue{the real world's} $P(f)$, or at least about how well
one or more search algorithms perform on \blue{that} $P(f)$.
For example, we could do this by recording
the results of running a particular search algorithm on a set of (randomly chosen) 
real-world search problems, and using
those results as a ``training set'' for a supervised machine learning algorithm
that models how those algorithms compare to one another on such search problems. The hope would be that by
doing this, we can give ourselves formal assurances that one
search algorithm should be used rather than another, for the $P(f)$ that
governs the real world.

The precise details of how well such an approach would perform depend on 
the precise way that it is formalized. However two broadly applicable
restrictions on its performance are
given by an inner product formula for supervised learning
and an associated NFL theorem for supervised learning.

\blue{Just like search, supervised learning involves an input space $X$,
an output space $Y$, a function $f$ relating the two, and a data set 
of $(x, y)$ pairs. The goal in supervised learning though is not to iteratively augment the 
data to find what $x$ minimizes the ``target function'' $f(x)$. Rather
it is  to take a fixed data set and estimate the entire function $f$.
Such a function mapping a data set to an estimate of $f$ (or more generally
an estimate of a distribution over $f$'s) is called a \textbf{learning algorithm}.
We then refer to the accuracy of the estimate for $x$'s that do not occur in
the data set as \textbf{off-training set error}. More precisely, in supervised learning
we are concerned with the expected value of a ``loss function'' over points
outside of the training set, which plays the same role as 
the performance measure $\Phi$ does in search.

The supervised learning inner product formula tells us that
the performance of any supervised learning algorithm is governed
by an inner product between two vectors, both indexed by the set
of all target functions. In particular, as long as the loss function is symmetric, it tells us that
how ``aligned'' the supervised  learning algorithm is with the real world
(i.e., with the posterior distribution of target functions conditioned on a training set) determines how well 
that algorithm will generalize from any training set to a separate test set. (See appendix.)}
This supervised learning inner product formula results in a set of NFL theorems for
supervised learning, applicable when some additional common conditions
concerning  the loss function hold. In some ways these theorems are even more striking
than the NFL for search theorems.

As an example,
let $\Theta$ be a set of the favorite supervised learning algorithms of some scientist $\A$. So when given a training set
$d$, scientist $\A$ estimates what $f$ produced that training set the
 following way. First they run cross-validation on $d$ to compare the algorithms in $\Theta$.
 They then choose the algorithm $\theta \in \Theta$ with lowest such cross-validation error. As a final step, they run that algorithm
 on all of $d$. In this way $\A$ generates their final hypothesis $h$ to generalize from $d$. 

Next suppose that scientist $\BB$ has the same set of favorite learning algorithms. So they 
decide how to generalize from a given data set the same way as $\A$ does --- but with a twist. For some reason, $\BB$ uses \emph{anti}-cross-validation
rather than cross-validation. So the
 the algorithm they choose to train on all of $d$ is the element of $\Theta$ with \emph{greatest} cross-validation
 error on $d$, not the one with the smallest such error. 

Note that since $\Theta$ is fixed, the procedure run by scientist $\A$ is simply a rule
that maps any arbitrary training set $d$  to an estimate of the target function for $x$ outside of that training set.
In other words, $\A$ themselves constitute a supervised learning algorithm. Similarly, since $\Theta$ is fixed,
$\BB$ is a (different) supervised learning algorithm.
So by the NFL theorems for supervised learning, we have no \emph{a priori}
 basis for preferring scientist $\A$'s hypothesis to scientist $\BB$'s. Although it is difficult to actually produce such $f$'s
 in which $\BB$ beats $\A$, by the NFL for supervised learning theorem
 we know that there must be ``as many'' of them (weighted by performance)
 as there are $f$'s for which $\A$ beats $\BB$. In other words, anti-cross-validation
beats cross-validation as often as the reverse. 
 
 Despite this lack of formal guarantees behind cross-validation in supervised learning, 
 it is hard to imagine any scientist who would not prefer 
 to use it to using anti-cross-validation. Indeed, one can view cross-validation 
 (or more generally ``out of sample'' techniques) as a formalization of the
 scientific method: choose among theories according to which better fits experimental data that
 was generated after the theory was formulated, and then use that theory to make
 predictions for new experiments. By the inner product formula for supervised learning,
 this bias of the scientific community  in favor of using out-of-sample techniques in general, and
 cross-validation in particular, must correspond somehow to a bias in favor of a particular $P(f)$.
 This implicit prior $P(f)$ is quite difficult to express mathematically. 
 Yet almost every conventional supervised learning
 prior (e.g., in favor of smooth targets) or non-Bayesian bias favoring some learning algorithms over others
(e.g., a bias in favor of having few degrees of freedom in a hypothesis class, 
in favor of generating a hypothesis with low
 algorithmic complexity, etc.) is often debated by members of the scientific community. In contrast, nobody debates
 the ``prior'' implicit in out-of-sample techniques. Indeed, it is exactly this prior which justifies
 the ubiquitous use of contests involving hidden test data sets to judge which of a set of learning
algorithms are best.

\section{Implications of NFL for other formal results concerning inference, and for philosophy of science }

It is worth taking a moment to describe how the NFL theorems can be reconciled with the 
many proofs in the literature of lower bounds on generalization error 
which would appear to provide \textit{a priori} reason to prefer one algorithm over another.
Briefly, there are two problematic aspects to those proofs. First, the NFL theorems all
concern the conditional distribution 
\eq{
P(\Phi \mid d, \A)
}
where $\Phi$ is the random variable giving the expected loss of the prediction made by 
algorithm $\A$ for test points outside of the training set data $d$. In particular, they tell us that
\eq{
\EEE(\Phi \mid d, \A) = \EEE(\Phi \mid d, \BB)
}
for any two algorithms $\A$ and $\BB$.

This equation concerns the posterior expected (off-training-set)
loss, conditioned on $d$ --- which according to Bayesian decision theory, should guide
our decision of which algorithm to use, $\A$ or $\BB$. One can average over $d$ (produced by sampling the implicit prior
$P(f)$), to see that NFL also tells us that
\eq{
\EEE(\Phi \mid m, \A)  = \EEE(\Phi \mid m, \BB)
}
the expected off-training set loss conditioned only on $m$, the size of the data set $d$, not the
precise data set. 

Next, let $\Phi'$ be the expected loss over the entire space $X$, not
just the portion of $X$ outside of $d_X$. There are many formal results in the literature which concern $\Phi'$, 
not $\Phi$. In addition, many of these results don't explicitly
specify what the conditioning event is for the distributions they calculate, simply writing
them without any conditioning event at all. However when you dig into the proofs,
you often find that the results concern conditional distributions like
\eq{
P(\Phi' \mid m, \A, f)
}
Note that in a formal sense, this conditional distribution is ``backwards'' --- it conditions
on $f$, which is what is  unknown, and averages over $d$'s, even though the actual $d$ \textit{is} known.
(This criticism of the choice of conditioning event is, of course, a central issue in the age-old controversy between Bayesian statistics and non-Bayesian statistics.)

Often these results are independent of $f$, which sometimes leads researchers 
to interpret them as meaning that some algorithm $\A$ should be used rather than some other
algorithm $\BB$, ``no matter what the distribution over $f$'s is''. In particular, the argument
is often made that so long as $m$ is far smaller than the size of the space $X$, $\Phi'$ will
approximate $\Phi$ arbitrarily well with arbitrarily high probability, and therefore a result
like $\EEE(\Phi' \mid m, \A, f) < \EEE(\Phi' \mid m, \BB, f) \; \forall f$
means that algorithm $\A$ is better than $\BB$ on off-training set error, no matter what $P(f)$ is. That reasoning
is simply wrong though; if one tries to use the standard rules of probability theory
to convert results like $\EEE(\Phi' \mid m, \A, f) < \EEE(\Phi' \mid m, \BB, f)$  to results
like $\EEE(\Phi \mid m, \A) < \EEE(\Phi \mid m, \BB)$
one fails (unless one makes an assumption for $P(f)$). This is true no matter how big $X$ is compared to $m$.
In fact, it is not trivial to make the transition from results concerning $\EEE(\Phi' \mid m, \A, f)$
to results concerning $\EEE(\Phi' \mid d, \A)$~\cite{wolp95b,wolpert1997bias}.

There has also been work  
that has claimed to derive a ``Bayesian Occam's
razor'' using Bayes factors, where the analysis is over (ultimately arbitrarily defined) \textit{models} of problems,
rather than over individual problems directly~\cite{mack03,jefferys1992ockham,lore90,gull88}. However, NFL tells us that such approaches must,
ultimately, simply be hiding their assumption concerning the prior over problems. Indeed, as \textit{reductio ad absurdum}, 
one can use the kind of reasoning promoted in these papers to imply that \textit{any} algorithm is superior to any other, by appropriately
redefining the models~\cite{wolpert1995bayesian}. 
There has also been work that claims to use algorithmic information
theory~\cite{livi08} to refute NFL~\cite{lattimore2013no}. However, ultimately this work simply makes an assumption for
a particular prior, and then shows that there are \textit{a priori} distinctions between algorithms for that prior
(in this case, the prior of algorithmic information theory, which explicitly
prefers hypotheses that can be encoded in shorter programs). Of course, this is completely consistent with NFL.
More broadly, other work has shown that all one needs to assume is that as one's algorithm is fed more and more data
it performs better (on average), in order to justify a particular form of Occam's razor~\cite{wolpert1990relationship}. This too is consistent with
NFL.

%

The implications of NFL for the entire scientific enterprise are also wide-ranging. In particular, we can let  
$X$ be the specification of how to configure an experimental apparatus, and $Y$
the outcome of the associated experiment. So $f$ is the relevant physical laws
determining the results of any such experiment, i.e., they are a specification of a universe. In addition, $d$ is a set of such experiments,
and the function $h$ produced by the ``learning algorithm'' is a theory that tries to explain that experimental data ($P(h \mid d)$
being the distribution that embodies the scientist who generates that theory).
Under this interpretation, off-training set error quantifies how well any theory produced by a 
particular scientist predicts the results of
experiments not yet conducted. So roughly speaking,
according to the NFL theorems for search,
if scientist $\A$ does a better job than scientist $\BB$ of producing accurate theories from data for one set of universes,
scientist $\BB$ will do a better job on the remaining set of universes. This is true
even if both universes produced the exact same set of scientific data that
the scientists use to construct their theories \blue{--- in which case it is theoretically impossible
for the scientists to use any of the experimental data they have ever seen \emph{in any way
whatsoever} to determine which set of universes they are in.}

As another implication of NFL for supervised learning, take
 $x \in X$ to be the specification of an objective function, and say we have
two professors, Smith and Jones, each of whom when given any such $x$ will
produce a search algorithm to run on $x$. Let $y \in Y$ be the bit that equals 1 iff
the performance of the search algorithm produced by Prof. Smith is better than the
performance of the search algorithm produced by Prof. Jones.{\footnote{\blue{Note that as a special case,
we could have each of the two professors always produce the exact same search algorithm
for any objective function they are presented.
In this case comparing the performance of the two professors just amounts
to comparing the performance of the two associated search algorithms.}}  So any training
set $d$ is a set of objective functions, together with the bit of which of 
(the search algorithms produced by) the two professors on those objective functions 
performed better. 

Next, let the learning algorithm $\C$ be the
simple rule that we predict $y$ for all $x \not \in d^m_X$ to be 1 iff the majority of the values in
$d^m_Y$ is 1, and the learning algorithm $\DDD$ to be the rule that we predict $y$ to be -1 iff the majority of the values in
$d^m_Y$ is 1. So $\C$ is saying that if Professor Smith's choice of search algorithm outperformed
the choice by Professor Jones the majority of
times in the past,  then predict that they will continue to outperform Professor Jones in the future. In contrast, $\DDD$ is
saying that there will be a magical flipping of relative performance, in which suddenly Professor
Jones is doing better in the future, if and only if they did worse in the past. 

The NFL for supervised
learning theorem tells us that there are as many universes in which algorithm $\C$
will perform worse than algorithm $\DDD$  --- so that Professor Jones magically starts
performing worse than Professor Smith --- as there are universes the other way around. This is true even if
Professor Jones produces the random search algorithm no matter what the value of $x$
(i.e., no matter what objective function they are searching).
In other words, just because Professor Smith produces search algorithms that outperform random
search in the past, without making some assumption about the probability distribution over
universes, we cannot conclude that they are likely to continue to do so in the future.

The possible implications for how tenure decisions and grant awards are made will not be considered here.

\section{Exploiting the relation between supervised learning and search to improve search}

\red{Given the preceding discussion, it seems that 
supervised learning is closely analogous to
search, if one replaces the ``search algorithm'' with a ``learning algorithm'' and 
the ``objective function'' with a ``target function". So it should not
be too surprising that the inner product formula and NFL theorem for search have analogs
in supervised learning. This close formal relationship between search
and supervised learning} means that techniques
developed in one field can often be ``translated'' to apply directly to the other field.

\red{A particularly  pronounced example of this occurs in the simplest (greedy) form of the Monte Carlo Optimization (MCO) approach
to search~\cite{erno98}. In that form of MCO,} one uses a data set $d$ to form a distribution $q(x \in X)$ rather than \blue{(as in
most conventional search algorithms)}
directly form a new $x$. That $q$ is chosen so that that one expects the expected value of the
objective function, $ \sum_x q(x) f(x)$ to have a low value, i.e., so that one expects
a sample of $q(.)$ to produce an $x$ with a good value of the objective function. One then forms a
sample $x$ of that $q(.)$, and evaluates $f(x)$. This provides a new pair $(x, f(x))$ that gets added
to the data set $d$, and the process repeats.

MCO algorithms can be viewed as
variants of random search algorithms like genetic algorithms and simulated
annealing, in which the random distribution governing which point 
to sample next is explicitly expressed and controlled, rather than be 
implicit and only manipulated indirectly. Several other algorithms can be cast as forms of MCO
(e.g., the cross-entropy method~\cite{rukr04}, the MIMIC algorithm~\cite{deis97}). 
MCO algorithms differ from one another in
how they form the distribution $q$ for what point next to sample, with some not trying
directly to optimize $ \sum_x q(x) f(x)$ but instead using some 
other optimization goal. 

It turns out that the problem of how best to 
choose a next $q$ in MCO is formally identical to the supervised learning problem of how best to choose a
hypothesis $h$ based on a training set $d$~\cite{rawo07,rawo08}. 
If one simply re-interprets all MCO variables 
as appropriate supervised learning variables, one transforms any MCO problem into a supervised learning problem
(and vice-versa).
The rule for this re-interpretation  is effectively a dictionary that allows us to transform any technique
that has been developed for supervised learning into a technique for (MCO-based) search. Regularization, bagging,
boosting, cross-validation, stacking, etc., can all be transformed this way into techniques to improve
search. 

As an illustration, 
we can use the dictionary to translate the use of cross-validation to choose  a hyperparameter from
the domain of supervised learning into the domain of search. 
Training sets become data sets, and the
hyperparameters of a supervised learning algorithm become the parameters of an MCO-based search algorithm.
For example, a regularization constant in supervised learning gets transformed into the temperature
parameter of  (a form of MCO that is a small variant of) simulated annealing. In this way 
using the dictionary to translate cross-validation into the search domain
shows us how to use it on one's data set in search to dynamically
update the temperature in the temperature-based MCO search algorithm. That updating proceeds by running the
MCO algorithm repeatedly on subsets of
one's \emph{already existing} data set $d$. (No new samples of the objective function $f$ 
beyond those already in $d$ are involved in this
use of cross-validation for search, just like no new samples are involved in
the use of cross-validation in supervised learning.)

Experimental tests of MCO search algorithms designed by using the dictionary have established that they work
quite well in practice~\cite{rawo07,rawo08}. Applying the dictionary to create analogs of bagging and and stacking in the context of search, 
in addition to creating analogs of cross-validation, have all
been found to transform an initially powerful search algorithm into a new one with improved search performance.

 Of course, these experimental results do not mean there is any formal justification for
 these kinds of MCO search algorithms; NFL for search cannot be circumvented.

\section{Free lunches and future research}

There are many avenues of research related to the NFL theorems which have not yet been
properly explored. Some of these involve \emph{free lunch}
theorems which concern fields closely related to search, e.g., co-evolution~\cite{woma05}. Other free lunches arise
in supervised learning, e.g., when the loss function does not obey the conditions that were alluded to above~\cite{wolpert1997bias}.

However it is important to realize that none of these (no) free lunch theorems concern the \emph{covariational} behavior
of search and / or learning algorithms. For example, despite the NFL for search theorems, there are scenarios
where, for some $f$'s, $\EEE(\Phi \mid f, m, \A) - \EEE(\Phi \mid f, m, \BB) = k$ (using the notation of
the appendix), but there are no $f$'s for which the reverse it true, i.e., for which the difference
$\EEE(\Phi \mid f, m, \BB) - \EEE(\Phi \mid f, m, \A) = k$. It is interesting to speculate that such ``head-to-head'' distinctions might ultimately
provide a rationale for using many almost universally applied heuristics, in particular for
using cross-validation rather than anti-cross-validation in both
supervised learning and search.

There are other results where, in contrast to the NFL for search theorem, one does not consider fixed search algorithms
and averages over $f$'s, but rather fixes $f$ and averages over algorithms. These results allow
us to compare how intrinsically hard it is to search over a particular $f$.
They do this by allowing us to compare two $f$'s based on the sizes of the sets of algorithms that do better than 
the random algorithm does on those $f$'s~\cite{mawo96}. While there are presumably analogous results for
supervised learning, which would allow us to measure how intrinsically hard it is to learn
a given $f$, nobody currently knows. All of these issues are the subject
of future research.


\section*{Appendix}
\addcontentsline{toc}{section}{Appendix}

\subsection*{A.1\;\;\;  NFL and inner product formulas for search}


To begin, expand the performance probability distribution:
\begin{eqnarray}
P(\phi \mid \A, m) &=& \sum_{d^m_Y} P(d^m_Y \mid \A, m) P(\phi \mid d^m_Y, \A, m) \nonumber \\
 &=& \sum_{d^m_Y} P(d^m_Y \mid \A, m) \delta(\phi, \Phi(d^Y_m))
 \label{eq:1}
\end{eqnarray}
where the delta function equals 1 if its two arguments are equal, zero otherwise.
%
The choice of search algorithm affects performance only through the term $P(d^m_Y \mid \A, m)$. 
In turn, this probability of $d^m_Y$ under $\A$ is given by
\begin{eqnarray}
P(d^m_Y \mid \A, m) &=& \sum_{f} P(d_Y^m \mid f, m, \A) P(f \mid m, \A) \nonumber \\
&=& \sum_{f} P(d_Y^m \mid f, m, \A) P(f).
\end{eqnarray}

Plugging in gives
\eq{
P(\phi \mid \A, m) &= \sum_{f} P(f) D(f; d^m_Y, \A, m) 
\label{eq:inner_product}
}
where
\eq{
D(f; d^m_, \A, m)  &:= \sum_{d^m_Y} P(d^m_Y \mid f, \A, m) \delta(\phi, \Phi(d^Y_m))
}
So for any fixed $\phi$, $P(\phi \mid \A, m)$ is an inner product of two real-valued vectors each indexed by $f$:
$D(f; d^m_Y, \A, m) $ and $P(f)$. Note that all the details
of how the search algorithm operates are embodied in the first of those vectors. In contrast,
the second one is completely independent of the search algorithm.

This notation also allows us to state the NFL for 
search theorem formally. Let
$B$ be any subset of the set of all objective functions, $Y^X$.
Then \cref{eq:inner_product} allows us to express the expected performance
for functions inside $B$ in terms of expected performance outside of $B$:
\eq{
\sum_{f \in B} \EEE(\Phi \mid f, m, \A) &= constant - \sum_{f \in Y^X \setminus B} \EEE(\Phi \mid f, m, \A)
\label{eq:nfl}
}
where
the constant on the right-hand side depends on the performance measure $\Phi(.)$, but is independent of both $\A$ and
$B$~\cite{woma97}.  Expressed differently, \cref{eq:nfl}
says that $\sum_f E(\Phi \mid f, m, \A)$ is independent of $\A$.
This is the core of the NFL for search, as elaborated in the next section.

\subsection*{A.2\;\;\; NFL for search when we average over $P(f)$'s}

To derive the NFL theorem that applies when we vary over $P(f)$'s, first recall
our simplifying assumption 
that both $X$ and $Y$ are finite (as they will be when doing search on any digital
computer).  Due to this, any $P(f)$ is a finite dimensional real-valued vector living on a simplex $\Omega$.
Let $\pi$ refer to a generic element of $\Omega$. So $\int_\Omega d\pi \; P(f \mid \pi)$
is the average probability of any one particular $f$, if one uniformly averages over all 
distributions on $f$'s. By symmetry, this integral must be a constant, independent of $f$.
In addition, as mentioned above, \cref{eq:nfl} tells us that $\sum_{f \in B} \EEE(\Phi \mid f, m, \A)$ is 
independent of $\A$. Therefore for any two search algorithms
$\A$ and $\BB$, 
\begin{eqnarray}
\sum_f \EEE(\Phi \mid f, m, \A) &=& 
   \sum_f \EEE(\Phi \mid f, m, \BB) ,\nonumber \\
\sum_f \EEE(\Phi \mid f, m, \A) \bigg[\int_\Omega d\pi \; P(f \mid \pi)\bigg] &=& 
   \sum_f \EEE(\Phi \mid f, m, \BB) \bigg[\int_\Omega d\pi \; P(f \mid \pi)\bigg] ,\nonumber \\
\sum_f \EEE(\Phi \mid f, m, \A) \bigg[\int_\Omega d\pi \; \pi(f)\bigg] &=& 
    \sum_f \EEE(\Phi \mid f, m, \BB) \bigg[\int_\Omega d\pi \; \pi(f)\bigg], \nonumber \\
\int_\Omega d\pi \; \sum_f \EEE(\Phi \mid f, m, \A) \pi(f) &=&  \int_\Omega d\pi \; \sum_f \EEE(\Phi \mid f, m, \BB) \pi(f),
\end{eqnarray}
i.e.,
\begin{eqnarray}
\int_\Omega d\pi \; \EEE_\pi(\Phi \mid m, \A) &=& \int_\Omega d\pi \; \EEE_\pi(\Phi \mid m, \BB).
\label{eq:5}
\end{eqnarray}
We can re-express this result as the statement that $\int_\Omega d\pi \; \EEE_\pi(\Phi \mid m, \A)$ is independent of $\A$.

Next, let $\Pi$ be any subset of  $\Omega$. Then our result that $\int_\Omega d\pi \; \EEE_\pi(\Phi \mid m, \A)$ is independent of $\A$ implies 
\begin{eqnarray}
\int_{\pi \in \Pi} d\pi \; \EEE_\pi(\Phi \mid m, \A) &=& constant - \int_{\pi \in \Omega \setminus \Pi} d\pi \; \EEE_\pi(\Phi \mid m, \A)
\label{eq:nfl_prior}
\end{eqnarray}
where the constant depends on $\Phi$, but is independent of both $\A$ and
$\Pi$. So if any search algorithm performs particularly well for one set of $P(f)$'s, $\Pi$,
it must perform correspondingly \emph{poorly} on all other $P(f)$'s. This is the NFL
theorem for search when $P(f)$'s vary.

\subsection*{A.3\;\;\; NFL and inner product formulas for supervised learning}

To state the supervised learning inner product and NFL theorems requires introducing
some more notation. Conventionally, these theorems
are presented in the version where both the the learning
algorithm and target function are stochastic. (In contrast, the restrictions for search ---
presented above --- conventionally involve a deterministic search algorithm and deterministic objective
function.) This makes the statement of the restrictions for supervised learning intrinsically more
complicated.

Let $X$ be a finite \textbf{input} space, $Y$ a finite \textbf{output} space, and say we have 
a \textbf{target distribution} $f(y_f \in Y \mid x \in X)$, along with 
a \textbf{training set}  $d = (d^m_X, d^m_Y)$  of $m$ pairs $\{(d^m_X(i) \in X, d^m_Y(i) \in Y)\}$, that
is stochastically generated according to a distribution $P(d \mid f)$ (conventionally called a
\textbf{likelihood}, or ``data-generation process").
Assume that based on $d$ we have a \textbf{hypothesis distribution} $h(y_h \in Y \mid x \in X)$.
(The creation of $h$ from $d$ --- specified \emph{in toto} by the distribution $P(h \mid d)$ ---
is conventionally called the \textbf{learning algorithm}.)
In addition, let $L(y_h, y_f)$ be a \textbf{loss function} taking $Y \times Y \rightarrow \R$.
Finally, let $C(f, h, d)$ be an \textbf{off-training set cost function}{\footnote{The choice to use an off-training set cost function 
for the analysis of supervised learning is the analog of the
choice in the analysis of search to use a search algorithm that
only searches over points not yet sampled. In both the cases, the goal is to ``mod out"
aspects of the problem that are typically not of interest and might result in misleading results: ability
of the learning algorithm to reproduce a training set in the case of supervised learning,
and ability to revisit points already sampled with a good objective value in the case of
search.}},
\begin{eqnarray}
C(f, h, d) \propto \sum_{y_f \in Y, y_h \in Y} \sum_{q \in X \setminus d^m_X} P(q) L(y_f, y_h) f(y_f \mid q) h(y_h \mid q)
\end{eqnarray}
where $P(q)$ is some probability distribution over $X$ assigning non-zero measure to $X \setminus d^m_X$.

All aspects of any supervised learning scenario --- including the prior, the learning algorithm, the
data likelihood function, etc. --- are given by the joint distribution $P(f, h, d, c)$ (where $c$ is values of the cost function)
and its marginals.
In particular, in ~\cite{wolp95b} it is proven that the probability of a particular cost value $c$ is given by
\begin{eqnarray}
P(c \mid d ) &=& \int df dh \; P(h \mid d)P(f \mid d)M_{c,d}(f, h)
\label{eq:ip_sup_learn}
\end{eqnarray}
for a matrix $M_{c, d}$ that is symmetric in its arguments so long as
the loss function is. $P(f \mid d) \propto P(d \mid f) P(f)$ is the posterior probability that the real world
has produced a target $f$ for you to try to learn, given that you only know $d$. It has nothing
to do with your learning algorithm. In contrast, $P(h \mid d)$ is the
specification of your learning algorithm. It has nothing to do with the distribution of
targets $f$ in the real world.
So \cref{eq:ip_sup_learn} tells us that as long as the loss function is symmetric,
how ``aligned" you (the learning algorithm) are with the real world
(the posterior) determines how well you will generalize.

This supervised learning inner product formula results in a set of NFL theorems for
supervised learning, once one imposes some additional conditions on the loss function.
See~\cite{wolp95b} for details.

\noindent \bibliographystyle{unsrt}
\bibliography{../../../../../BIB/refs}

\end{document}